# Extraction of semantic relations from a Basque monolingual dictionary using Constraint Grammar

Eneko Agirre, Olatz Ansa, Xabier Arregi, Xabier Artola, Arantza Díaz de Ilarraza,

Mikel Lersundi, David Martínez, Kepa Sarasola, Ruben Urizar

Computer Science Faculty – University of the Basque Country
P.O. box 649, E-20080 Donostia
Basque Country
e-mail: eneko@si.ehu.es

**ABSTRACT**

This paper deals with the exploitation of dictionaries for the semi-automatic construction of lexicons and lexical knowledge bases. The final goal of our research is to enrich the Basque Lexical Database with semantic information such as senses, definitions, semantic relations, etc., extracted from a Basque monolingual dictionary. The work here presented focuses on the extraction of the semantic relations that best characterise the headword, that is, those of synonymy, antonymy, hypernymy, and other relations marked by specific relators and derivation. All nominal, verbal and adjectival entries were treated. Basque uses morphological inflection to mark case, and therefore semantic relations have to be inferred from suffixes rather than from prepositions. Our approach combines a morphological analyser and surface syntax parsing (based on Constraint Grammar), and has proven very successful for highly inflected languages such as Basque. Both the effort to write the rules and the actual processing time of the dictionary have been very low. At present we have extracted 42,533 relations, leaving only 2,943 (9%) definitions without any extracted relation. The error rate is extremely low, as only 2.2% of the extracted relations are wrong.

## 1. Introduction

This paper deals with the exploitation of dictionaries for the semi-automatic construction of lexicons and lexical knowledge bases [AMSLER, 1981; CALZOLARI/PICCHI, 1986; BOGURAEV/BRISCOE, 1989; RICHARDSON ET AL., 1998]. Our research group has been previously involved in creating a Dictionary Knowledge Base from the definitions in a French monolingual dictionary [ARTOLA, 1993; AGIRRE ET AL., 1997]. The final goal of the present research is to enrich the Basque Lexical Database (EDBL) [ADURIZ, 1998] with semantic information such as word senses, definitions, semantic relations, etc., extracted from a Basque monolingual dictionary called *Euskal Hiztegia* [SARASOLA, 1996]. Previous to any linguistic processing, the structure of the dictionary was parsed and encoded following the Text Encoding Initiative guidelines [SPERBERG-MCQUEEN/BURNARD, 1994]. The definitions and examples of the dictionary have been parsed using Constraint Grammar [KARLSSON ET AL., 1995; TAPANAINEN, 1996], which was also used to extract the semantic relations.

The work here presented focuses on the extraction of the semantic relations that best characterise the headword, that is, those of synonymy, antonymy, hypernymy, and other relations marked by specific relators and derivation. All nominal, verbal and adjectival entries were treated. It must be highlighted that Basque uses morphological inflection to mark case, which means that semantic relations have to be inferred from suffixes rather than from prepositions.

In the following section we present the target dictionary and the relations we have been seeking. In section 3 the method used to extract the relations and the results obtained are shown. The mapping from surface relations to semantic relations is discussed in section 4. Finally some conclusions are drawn.

**2.  Features of the dictionary**

*Euskal Hiztegia* [SARASOLA, 1996] is a monolingual dictionary of Basque. It is normative and repository of standard Basque. It was produced based mainly on literary tradition. The dictionary has 30,715 entries and 41,699 senses.

The source dictionary was originally in Rich Text Format, lacking any structure except typographical codes. In order to structure the entries and identify the fields[1], the dictionary was parsed using a Definite Clause Grammar. The structured entries are encoded in SGML following the TEI guidelines for monolingual dictionaries [SPERBERG-MCQUEEN/BURNARD, 1994]. The whole process of conversion and the application of the TEI representation are covered in [ARRIOLA ET AL., 1995; ARRIOLA/SOROA, 1996].

The parsing of typographical codes into a structured representation is a painstaking process. At present 77% of the entries have been completely analysed without error, and an additional 20% of the entries are basically correct in regard to the structure of the entry, that is, the parts of speech, the sense numbers, the definitions and the examples. This yields 97% of the entries correct except errors in the date, in the grammar codes, or some other minor errors. For the task at hand the results of the parsing are highly satisfactory, but we are also planning to produce and release a hand-checked commercial version of the dictionary.

## 3. Superficial Relations Treated

According to Smith and Maxwell (1980), there are basically three ways to define a lexical entry:

- By means of a synonym, giving a word that has the same meaning. The headword and the synonym belong to the same part of speech.
- By means of a classic description, which follows the *genus et differentia specifica* pattern. The meaning of the headword is given by a generic term (the genus) and the description of the features that distinguish the headword (*differentia*) from the given generic term. The genus is usually the core of the definition sentence. The genus and the headword usually belong to the same part of speech. A hypernymy relation holds between the headword and the genus, that is, the genus is a generic term, and the headword is a more specific term.
- By means of specific relators, which are specialised syntactic ways of linking a definition word with the headword. The core of the definition and the headword are not necessarily of the same part of speech. The specific relator used in the definition often determines the semantic relation that holds between the headword and the core of the definition.

A single definition can contain all of the three defining patterns, some of them repeatedly. In addition, the headword can be a derived form. In this case the suffix or prefix will yield the relation between the headword and the root [PENTHEDOURAKIS/VANDERWENDE, 1993]. Some examples are shown below:

- "akabatu. Bukatu *(syn.)*, amaitu *(syn.)*."[2]
- "aireontzi. Hegalda daitekeen *(differentia)* zernahi ibilgailu *(genus / hipernym)*."[3]
- "ezpara. Habea *(syn.)*, Tabanidae familiako intsektuei *(related word)* ematen zaien izena *(relator)*."[4]
- "alaitsu *(derivation)*. Alaitasunez betea."[5]

## 4. Method

The first step is the tokenisation and the morphosyntactic analysis and tagging of the definition units. Basque is an agglutinative language with high morphologic complexity, which makes robust morphosyntactic analysis very essential [ADURIZ ET AL., 1999]. We use MORFEUS, a robust morphosyntactic analyser for Basque [URKIA, 1997; ALEGRIA ET AL., 1996].

In a second step, the structure of the definition is analysed in order to locate the definition patterns. As definitions can be very long and use awkward syntax, we do not try at present to analyse the whole definition, but we rather focus on the syntax around the distinctive patterns related to synonymy, "genus+differentia" and specific relators. Considering that the patterns change depending on the part of speech, we analyse separately nouns, verbs and adjectives.

In order to find out the distinctive patterns used by the lexicographer, we selected at random a sample of 200 definitions per part of speech. The patterns found are coded as mapping rules of Constraint Grammar [KARLSON ET AL., 1995; TAPANAINEN, 1996]. We parse the sample, evaluate the results and rewrite the rules in an iterative fashion. Once we get satisfactory results for the current sample, we select a new random sample and continue to write new rules and/or change the old ones. When the desired quality is obtained we stop, and use a last sample in order to evaluate the goodness of the present rules.

As a consequence of the application of the rules, the labels for the synonyms and the genera are added to the definition. In the case of the specific relators we both mark the word coding the relation (the *relator*) and the word which is related to the headword via the relator (the *related term*). Due to suffixes, both the relator and the related term are sometimes found in the same word.

As an example, we show below the output of the system for the definition of "gibelzorrotz".

"gibelzorrotz. Udarearen antzeko sagar *(related term)* mota *(relator)*."[6]

```
/<@@headword       gibelzorrotz>/<ID>/
/<@@POS     noun>/<ID>/
"<Udarearen>"
    "udare"   IZE ARR DEK GEN NUMS MUGM DEK ABS MG HAS_MAI DEF_HASI
"<antzeko>"
    "antzeko"  ADJ IZL DEK ABS MG
"<sagar>"
    "sagar"   IZE ARR  ZERO NOTGELGEN S:504 &ERLZ-MOTA10
"<mota>"
    "mota"  IZE ARR DEK ABS NUMS MUGM AORG NOTGELGEN S:497 &ERLT-MOTA
"<$.>"
     PUNT_PUNT
```

| Surface relator | Relation type | Part of speech of headword | Part of speech of related term |
| --- | --- | --- | --- |
| "-etako bat / -etako bakoitza" | Member of Taxonomy | noun | noun |
| "-en txikigarria" | Graduation | noun | noun |
| "-en handigarria" | Graduation | noun | noun |
| "-en kidea" | Synonymy Near synonymy Taxonomy | noun | noun |
| "-ri eman izena" | Taxonomy Hipernymy | noun | noun |
| "… mota" | Type of Taxonomy Hipernymy | noun | noun |
| "-era / -era bat" | Manner of Taxonomy | noun | noun |
| "… modukoa" | Near synonym | noun | noun |
| "-z mintzatuz" | Semantic field | noun | noun |
| "-ri dagokiona" | Corresponding to | noun | noun |
| "egin" | Product of | noun | noun |
| Adjective synonym pattern | Near synonymy | noun | adjective |
| Ellipsis | Role Possession | noun | verb |
| "-z betea" | Graduation | adjective | noun |
| "-ri dagokiona" | Corresponding to | adjective | noun |
| "-etako bat / -etako bakoitza" | Member of Taxonomy | adjective | noun |
| "-en txikigarria" | Graduation | adjective | adjective |
| "-en handigarria" | Graduation | adjective | adjective |
| "-ri eman izena" | Hypernymy | adjective | noun |
| Ellipsis | Role Possession | adjective | verb |
| "-z mintzatuz" | Semantic field | verb-noun | noun |

Table 1. List of relators. For some surface relators there is more than one possible relation type.

In the case of nouns we have 23 rules for synonyms, 37 for genera, and 26 for relators and related terms. For verbs we have 12 rules to mark synonyms, 22 for genera, and 1 for relators and related terms. Adjectives involve 5 rules to mark synonyms, 4 for genera, and 31 for relators and related terms. Table 1 shows the list of the relators and *deep* semantic relations that we are currently considering.

As an example we show below a couple of rules that cover a relator and a related term for nouns:[7]

```
MAP (&ERLT-MOTA) TARGET MOTA
        IF (-1 IZE-ZERO-NOTGELGEN) (1 PUNT/PKOMA/KOMA/DEF_BUKA) ;

MAP (&ERLZ-MOTA10) TARGET IZE-ZERO-NOTGELGEN
        IF (1 MOTA) (2 PUNT/PKOMA/KOMA/DEF_BUKA) ;
```

In addition to the analysis of the definitions, we tried to find derivational relationships. All nominal, adjectival and verbal headwords were morphologically analysed, trying to identify derivational morphemes. For instance, "alaitsu" (cf. endnote 5) can be analysed as "alai + tsu", and a relation between "alai" (joy) and "alaitsu" (joyful) can be established.

At present we have treated 8 nominal suffixes, 3 adjectival suffixes, 1 verbal suffix and 1 verbal prefix. The roots can be nouns, adjectives or verbs in all combinations, except verbal morphemes, which have always a verbal root.

## 5. Results

In order to evaluate the results we took three sets of 100 previously unseen random definitions, one for each part of speech.

### *5.1. Nouns*

We first identified by hand all synonyms, genera and specific relators present in the sample (*Target* column in table 2). After tagging the sample we evaluated how many words were correctly labelled (*OK* column), how many were incorrectly labelled (*Wrong* column), how many were marked (*Marked* column, which equals *OK* plus *Wrong*), and how many were missed (*Missed* column). The coverage indicates how many of the target semantic relations were actually found (*OK* divided by *Target*) and the error rate indicates how many were incorrectly labelled (*Wrong* divided by *Marked*).

Overall, the coverage on all target semantic relations for the nouns in the sample is 93.4%, and the error rate is 2.8%. Many definitions have more than one labelled word, giving an average of 1.5 semantic relations per definition. From the 100 definitions in the sample, we were able to find at least one semantic relation in 97 of them (*Definitions* row in table 2).

|         | Target | OK  | Wrong | Marked | Missed | Coverage (%) | Error rate (%) |
|---------|--------|-----|-------|--------|--------|--------------|----------------|
| **SYN** | 72     | 66  | 1     | 67     | 6      | 91.7         | 1.5            |
| **GEN** | 57     | 53  | 3     | 56     | 4      | 93.0         | 5.4            |
| **Relator** | 22 | 22  |       | 22     |        | 100.0        | 0.0            |
| **Overall** | 151 | 141 | 4    | 145    | 10     | 93.4         | 2.8            |
| **Definitions** | 100 | 97 |   | 97     | 3      | 97.0         |                |

Table 2. Results for the sample of nouns.

After studying the sample we proceeded to label all noun definitions in the dictionary (21,521) and we were able to identify at least one semantic relation in 93.7% of them, finding 1.49 semantic relations per definition.

### 5.2. Adjectives

In the adjectives the coverage for all relations in the sample is 90.8% (see *Overall* row in table 3), and the error rate is only 0.9%. We are able to find semantic relations for 77% of the definitions in the sample. The reason for the low coverage is that many definitions lack any recognisable pattern. Apparently the definition of adjectives has not been systematised in this dictionary.

|         | Target | OK  | Wrong | Marked | Missed | Coverage (%) | Error rate (%) |
|---------|--------|-----|-------|--------|--------|--------------|----------------|
| **SYN** | 42     | 37  |       | 37     | 5      | 88.1         | 0.0            |
| **GEN** | 8      | 8   | 1     | 9      |        | 100.0        | 11.1           |
| **Relator** | 70 | 64  |       | 64     | 6      | 91.4         | 0.0            |
| **Overall** | 120 | 109 | 1    | 110    | 11     | 90.8         | 0.9            |
| **Definitions** | 100 | 77 |   | 77     | 23     | 77.0         |                |

Table 3. Results for sample of adjectives.

In the dictionary there are 4,308 adjective definitions, and we mark at least one word in 3,162 of them (73.4% of the adjectival definitions), finding 1.48 semantic relations per definition.

### 5.3. Verbs

The coverage in the verb sample is 93.6% (see table 4), and the error rate is only 0.8%. We were able to find at least one relation for 92% of the definitions.

|             | Target | OK  | Wrong | Marked | Missed | Coverage (%) | Error rate (%) |
|-------------|--------|-----|-------|--------|--------|--------------|----------------|
| **SYN**     | 60     | 57  |       | 57     | 3      | 95.0         | 0.0            |
| **GEN**     | 78     | 72  | 1     | 73     | 6      | 92.3         | 1.4            |
| **Relator** | 2      | 2   |       | 2      |        | 100.0        | 0.0            |
| **Overall** | 140    | 131 | 1     | 132    | 9      | 93.6         | 0.8            |
| **Definitions** | 100 | 92 |       | 92     | 8      | 92.0         |                |

Table 4. Results for sample of verbs

In the dictionary there are 5,686 verb definitions, and we mark at least one word in 5,243 of them, 92.2% of definitions, finding 1.47 semantic relations per definition.

*5.4. Derivation*

To evaluate this approach we took a random sample of 100 derived headwords (see table 5). The morphological analyser has proved to have a very low error ratio but many derived headwords were missed, due to some limitations in our lexical database. We are currently extending the number of derivational suffixes in the database, focusing specially in adjectival suffixes, in order to compensate the low coverage in the analysis of the adjective definitions.

|                | Target | OK | Wrong |
|----------------|--------|----|-------|
| **Nouns**      | 35     | 13 | 1     |
| **Adjectives** | 49     | 26 |       |
| **Verbs**      | 8      | 8  |       |
| **Overall**    | 92     | 47 | 1     |

Table 5. Results of derivation.

**6. Contribution from a lexicographic point of view**

From a lexicographic point of view, we must underline that this research can contribute in different ways, giving us a different view of the dictionary making process and of the dictionary itself conceived as a tool "that explains words". This kind of work makes clear the dictionary viewed as a set of related senses or concepts, providing the user with more sophisticated ways of finding information when consulting it.

If we search for a word in an ordinary dictionary, we can know what it means and, we may find some words related with the entry. The research here presented can be used in the future for searching all the related words of a lexical entry, allowing the reader finding words he or she doesn't know or can't remember. So, the user will be able to find, for instance, all the synonyms of a given word, its generic term, the words that have some other relations with it, and its semantic group.

For example, we can look up the noun "madariondo" (Basque for "pear tree") in the *Euskal Hiztegia*. This definition tells us simply that it is a synonym of "udareondo", but if we want more information, we will have to look for "udareondo" in order to find out that it belongs to the rose family, and to know that its flowers are white. In our future tool all this information will be together and related.

The case of derivation shows us another example of related words that can be explicitly joined thanks to this kind of research. As a result, a big amount of words belonging to different parts of speech will be related.

## 7. Conclusions and future work

We have presented an efficient framework for the extraction of semantic relations from monolingual dictionaries. At present we focus on the main defining relations i.e. those of synonymy and hypernymy, and those conveyed by specific relators. These semantic relations will be stored in the lexical database for Basque. We have extracted 42,533 relations, leaving only 2,943 (9%) definitions, mostly adjectives, without any extracted relation. The error rate is very low, as only 2.2% of the extracted relations are wrong.

Our approach combines a morphological analyser and surface syntax parsing based on Constraint Grammar, which has proven very successful for a highly inflected language such as Basque. Both the effort to write the rules and the actual processing time of the dictionary have been very low.

In the future we plan to cover the semantic relations in the rest of the definition, that is, those relations involved in the part of the definition which is not the main defining pattern. For this we will be using more powerful partial parsers [ALDEZABAL ET AL., 1999]. Besides, the coverage of derivational phenomena is also being extended, focusing specially in adjectival suffixes, in order to reduce the number of adjectives without any relation.

In order to include the extracted relations in the lexical database, it is necessary to perform two disambiguation processes. On the one hand, there are some cases in which the surface relation extracted is ambiguous, that is, it could convey more than one deep semantic relation (cf. table 1). On the other hand, the word senses of the words in the semantic relation have to be also determined.

## 8. Acknowledgements

This work has been supported by the Department of Education, Universities and Research of the Basque Government, the University of the Basque Country (UPV-19/1999), the Government of Gipuzkoa under the Berbasare project (OF319-99), the European Fund of Development FEDER (2FD1997-1503), and the incentives given by the Gobernment of the Basque Country to High Performance Teams.

---

[1] The entries of the dictionary comprise the following: headword, part of speech, date information, sense grouping and number, definition, examples, subentries, etc. Most of the fields are optional, and can be repeated.

[2] Finish. Terminate *(syn.)*, complete *(syn.)*.

[3] Aircraft. Vehicle *(genus / hypernym)* which can fly *(differentia)*.

[4] Horsefly. Any deerfly, name given to *(relator)* some Tabanidae family insects *(related term)*.

[5] Joyful. Full of joy (the root "joy" is related to "joyful").

[6] Gibelzorrotz. Kind of *(relator)* apple *(related term)* similar to a pear. The relator is tagged with &ERLT-MOTA and the related term with &ERLZ-MOTA10 (10 is used for rule identification in debugging).

[7] The first rule could be paraphrased as follows: tag the word "mota" as being a relator (&ERLT-MOTA) if it is preceded by a noun in non-genitive form, and followed by a punctuation sign.

The second rule tags as related term (&ERLZ-MOTA) a non-genitive noun which is followed by the word "mota" and a punctuation sign.